# Effectiveness of Deep Networks in NLP using BiDAF as an example architecture

**Soumyendu Sarkar**
Stanford Center for Professional Development
soumyednu@gmail.com

## Abstract

Question Answering with NLP has progressed through the evolution of advanced model architectures like BERT and BiDAF and earlier word, character and context based embeddings. As BERT has leapfrogged the accuracy of models, an element of the next frontier can be the introduction of the deep networks and an effective way to train them. In this context I explored the effectiveness of deep networks focussing on the model encoder layer of BiDAF. BiDAF with its heterogeneous layers, provides the opportunity not only to explore the effectiveness of deep networks, but also to evaluate whether the refinements made in lower layers are additive to the refinements made in the upper layers of model architecture. I believe the next greatest model in NLP will in fact fold in a solid language modelling like BERT with a composite architecture which will bring in refinements in addition to generic language modelling, and will have a more extensive layered architecture. I experimented with the Bypass network, Residual Highway network, and DenseNet architectures. In addition I evaluated the effectiveness of ensembling the last few layers of the network. I also studied the difference character embeddings make in adding them to the word embeddings, and whether the effects are additive with deep networks. My studies indicate that deep networks are in fact effective in giving a boost. Also the refinements in the lower layers like embeddings are passed on additively to the gains made through deep networks.

## 1 Introduction

With the big steps of discovery of highly consequential NLP models like BERT, there have been incremental refinement and tuning of those base models for optimization of performance metrics, computation and memory, and adaptation to specific domains to suit business needs. These have resulted in architectures like ALBERT, XLNet, RoBERT and DistilBERT.

Exploring these optimizations involve steps in experimenting with model architectures, optimizations in model implementations, pre-processing and post-processing.

**One area of promise is using Residual Networks, Gated Highway Networks and DenseNet architectures**, which have been key in extending the computer vision CNN networks to deep networks while helping them train.

Even though BiDAF is a relatively earlier architecture, its simple stacked structure makes it ideally suited to experiment with extending the model encoder, with architectures like residual network or highway network and in model tuning like input embeddings and output layer. Some of **these learning can be carried forward to other model architectures as well.**

The **goal of this paper is to experiment with extension of the architecture of BiDAF with residual / highway networks and other architectural changes, and fine tuning the model, and evaluate the effect of these changes**.

This paper would explore whether refinements at these different layers are additive. In this context, we will start with the word embedding of Glove and then add CNN based character embedding, and measure the gains, both standalone and individually.

## 2 Prior Work

BiDAF paper from "*Minjoon Seo, Aniruddha Kembhavi, Ali Farhadi, and Hannaneh Hajishirzi. Bidirectional attention flow for machine comprehension*" provides key advancement of the neural attention mechanism prevalent before its publication. It introduced bi-directionality over uni-directional attention, and overcame the weakness of the attention weights at a given time step being function of the attended vector at a previous time step. Also it overcame the limitation of fixed size context representations. The most significant idea of the BiDAF is it's attention flow layer where attention flows both ways from the context to question and from the question to context.

I like the hierarchical layered architecture where it is layered like a process pipeline with the embedding layers, attention flow layer, modeling layer and the output layer. Even the embedding is layered with the modular embedding layers of character level, word level and bi-directional LSTM based contextual embeddings. This leaves a lot of room for innovation.

The bi-directional attention flow to obtain a query-aware context representation, brings in the advantage of letting the attention at every time step, along with the representations from previous layers, flow through to the modeling layer above. This is huge improvement over earlier fixed sized context representation summary.

Also the memory less attention mechanism makes the attention at each time step a function of just the query and the context paragraph at that time step and does not directly depend on the attention at the previous time step. This helps in divide and conquer of learning attention between the attention layer and the modelling layer. This helps the attention layer to focus on learning the attention between the query and the context. It also helps the modeling layer to focus on learning the interaction within the query-aware context representation. It also stops error propagation of incorrect attendances of previous time step to current time steps.

Also in BiDAF the attention flows both ways from the context to question and from the question to context. This was done with a bi-directional attention mechanisms for query-to-context and context-to-query.

These refinements helped BiDAF to go to the top of the leaderboard of the original SQUAD.

Also it's simple layered modular architecture is extendable enough to offer scope to add and experiment with different refinements.

The Highway Networks paper from "*Rupesh Kumar Srivastava, Klaus Greff, and Jürgen Schmidhuber. Highway networks*", helps us gain insight on how models can be optimized and trained for increasing depth, by allowing information flow across several layers with gating units which learn to regulate the flow.
Even though the depth of neural networks can often boost performance, network training through back-propagation becomes more difficult with increasing depth impeding performance. The Residual Networks proposes an architecture that allows information flow across several layers on "information highways", while introducing the use of gating units which learn to regulate the flow of information through a network. This eases gradient-based training of very deep networks.

The ability to train extremely deep networks opens up the possibility of studying the impact of depth on complex problems even in NLP with less restrictions. Various activation functions which may be more suitable for particular problems but for which robust initialization schemes are unavailable for vanishing gradients can be used in deep highway networks.

Highway Network essentially builds on the ResNet in a pretty intuitive way. The Highway Network preserves the shortcuts introduced in the ResNet, but augments them with a learnable parameter to determine to what extent each layer should be a skip connection or a nonlinear connection. This serves at the switch to determine to what extent information should be sent through the primary pathway or the skip pathway.

Finally the Dense Network, or DenseNet is an architecture that takes the insights of the skip connection to the extreme. Simplistically put, the core idea is of extending the skip connection from the previous layer, to connect every layer to every other layer. That way there is always a direct route for the information backwards through the network.

**Evaluating the feasibility of the deep networks** on a simpler structured architecture like BiDAF, and Exploring the limits can throw insights on how the similar enhancements can be carried forward to more complex models (like BERT ?).

Also evaluating the gains between the more basic pre-processing like word only embedding, or an **additive character** and word embedding throws insights on gains at different layers of a NLP model, which can also be carried forward to other model architectures.

## 3   Data

This project will be using Stanford's SQuAD 2.0 dataset in part. "Stanford Question Answering Dataset (SQuAD) is a reading comprehension dataset, consisting of questions posed by crowd workers on a set of Wikipedia articles, where the answer to every question is a segment of text, or span, from the corresponding reading passage, or the question might be unanswerable.
SQuAD2.0 combines the 100,000 questions in SQuAD1.1 with over 50,000 unanswerable questions written adversarially by crowd workers to look similar to answerable ones. To do well on SQuAD2.0, systems must not only answer questions when possible, but also determine when no answer is supported by the paragraph and abstain from answering."

We will be splitting the publicly available train and dev data sets into three splits: train, dev and test. To keep the train, dev and test data sets objective with knows matrix and vetted strategy for a good reference, in this project I will be using the Christopher Manning supervised CS224n project splits,
-       train (129,941 examples): All taken from the official SQuAD 2.0 training set.
-       dev (6078 examples): Roughly half of the official dev set, randomly selected.
-       test (5915 examples): The remaining examples from the official dev set, plus hand-labeled

I want to acknowledge Christopher Manning and his TAs for providing a baseline code for BiDAF with no character level embeddings.

To evaluate my model refinements, I will be quantifying the incremental gains or losses, and will be looking at both the final scores and the progression of scores with epochs to evaluate how expensive it is in achieving a desired performance level with training.

I will be evaluating the EM (Exact Match) and F1 scores which are the official SQuAD evaluation metrics, to measure the classification accuracy of my model for the generated answer.

Exact Match (EM) = (exactly matching answers) / (total evaluated answers)
 Note that the questions often have more than one answer provided and use a slightly more complex metrics.

F1 Score = 2 x Precision x Recall / (Precision + Recall),

I will be also evaluating the Answer vs. No Answer (AvNA), to measure its accuracy for no-answer predictions.

I will be leveraging the SQuAD 2.0 provided evaluation script used for the SQuAD 2.0 official evaluation, along with a sample prediction file that the script will take as input

## 4   Model

Baseline Model :
A simple Bidirectional Attention Flow (BiDAF) model with GloVe as word embedding and no character-level word embeddings is chosen as the baseline. This model did well with SQuAD 1,1, but was outperformed by BERT. But I like the hierarchical layered architecture where it is layered like a process pipeline with the Embedding layers, Attention flow layer, Modeling layer and the Output layer. This leaves a lot of room for innovation in the Embedding layer, Modeling layer, the Output layer and overall.

My inspiration has been ALBERT, and how ALBERT with its two parameter reduction techniques and a self-supervised loss for sentence-order prediction (SOP). is able to scale up to much larger configurations with fewer parameters than BERT-large, but achieve significantly better performance.

In contrast to that the Highway Networks paper helped me gain insight on how models can be optimized and trained for increasing depth, by allowing information flow across several layers with gating units which learn to regulate the flow.

In this paper we use three type of Deep Network techniques to extend the BiDAF architecture with additional model encoder layers with RNN encoder and Bidirectional Attention Flow with gates.
1. Simple Bypass Network
2. Highway Network
3. DenseNet Architecture

We evaluate the feasibility of this technique. We explore the limits of applying the highway network to BiDAF like architectures.

I also evaluate the performance gains in adding the character level word embeddings to the GloVe word embeddings of the baseline.

The models that I experimented with are the following:

### 4.1 BiDAF architecture with Bypass Networks

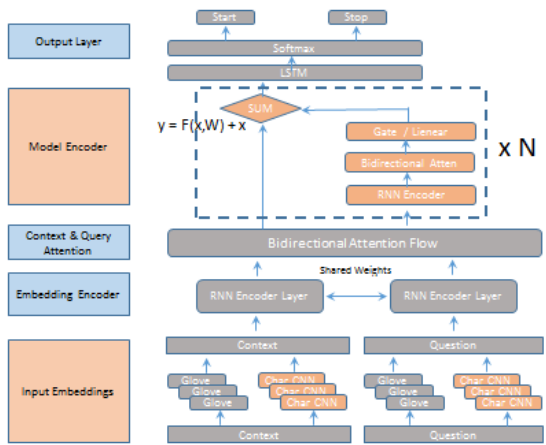

This has a simple ByPass architecture with the Model Encoder repeated in a deep network with a simple "y = F(x,W) + x" bypass.

### 4.2 BiDAF with Deep Highway Network

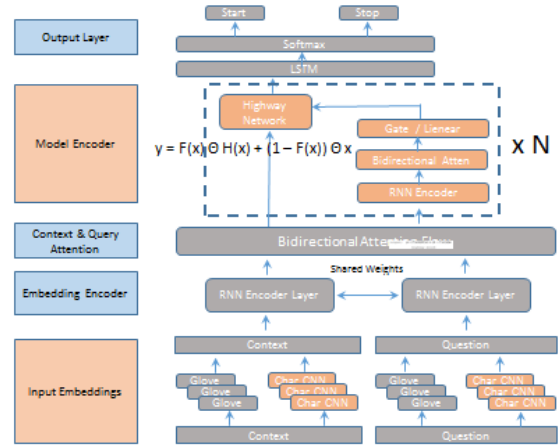

This has a Highway Network architecture with the Model Encoder repeated in a deep network with a transformation of "y = F(x) Θ H(x) + (1 − F(x)) Θ x."

### 4.3 BiDAF with DenseNet

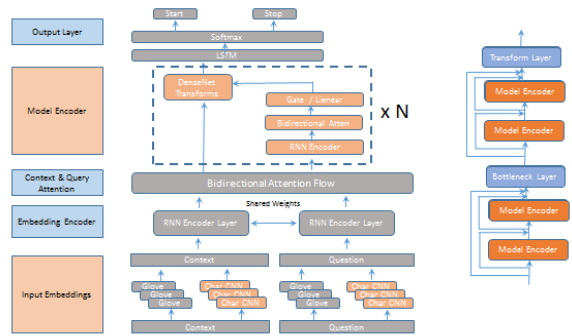

This has a DenseNet architecture with the Model Encoder repeated in sets of two DenseNet blocks with Bottleneck layers and Transform layers .
.

### 4.4 BiDAF with Ensembles of lower layers as output using Highway Network

I also extended the Highway Network model using an ensemble of lower layers as output going to the input to the LSTM encoder of the BiDAF output layer.

## 5 Results

The results are all captured with the Tensorboard evaluation graphics.

## 5.1 BiDAF architecture with Bypass Networks

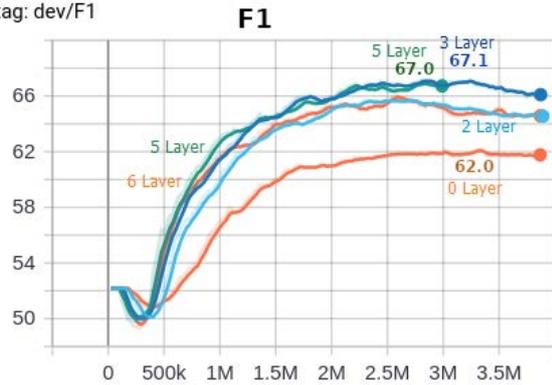

The Bypass network for model encoder boosted the F1 score from a baseline of 62.0, for no additional model network, to 67.1 for a 3 layer Bypass Network of model encoders. After a depth of 3 layers, the performance degrades for the simple bypass network unlike the Highway network, presented later.

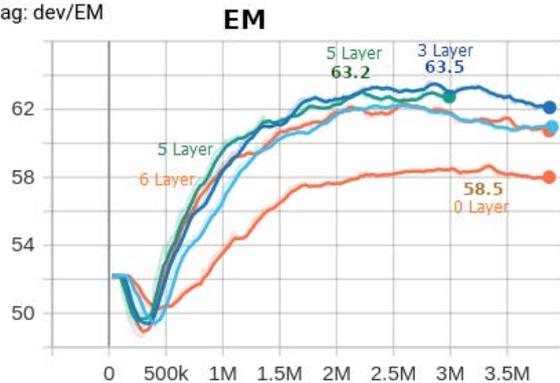

The Bypass network for model encoder boosted the EM score from a baseline of 58.5, for no additional model network, to 63.5 for a 3 layer Bypass Network of model encoders.

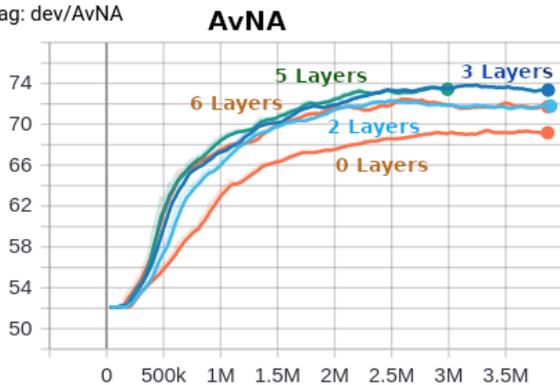

,

We observe similar improvements for Answer / No Answer with Bypass Layers.

As can be seen the 3 Layer deep network performs the best for Bypass network with a boost to both F1 score and EM.

## 5.2 BiDAF with Deep Highway Network

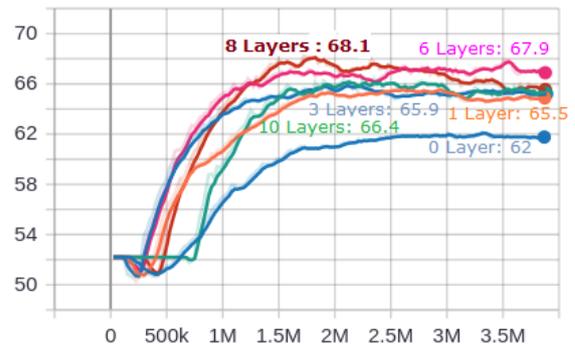

The 8 Layer deep network performs the best for Highway Network of model encoder with a boost of F1 score from 62.0 for baseline model encoder to 68.1. I also observe that after **8** layers, the performance goes down.

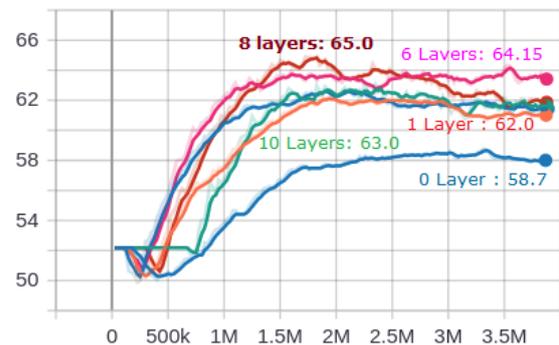

The **8** Layer deep network performs the best for Highway Network of model encoder with a boost of EM score from 58.7 for baseline model encoder, to 65.0.

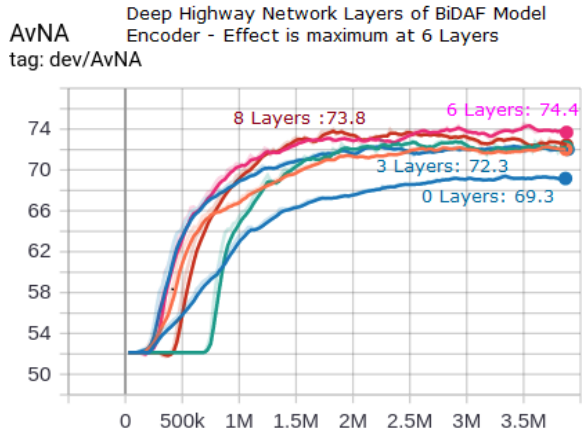

For Answer / No Answer, the **6** Layer network performs the best with a boost of F1 score from a 69.3 baseline model encoder to 74.4.

### 5.3   BiDAF with DenseNet

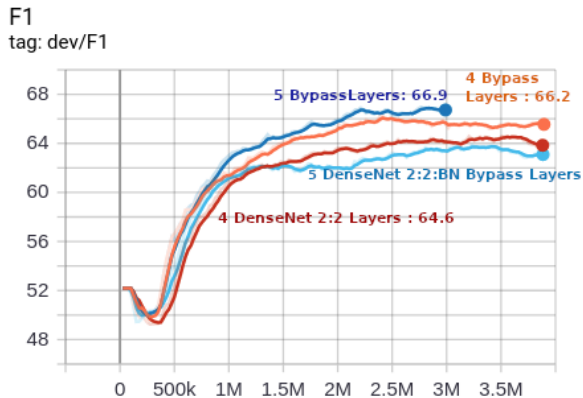

The DenseNet performed poorer than deep Bypass Networks of the BiDAF model encoder of same depth. I experimented with DenseNet blocks of 2+2, 3+3, and 2+2+1 with a bottleneck layers and transform layers. This needs to be further explored.

### 5.4   BiDAF with Output Layer Ensembles using Highway Network

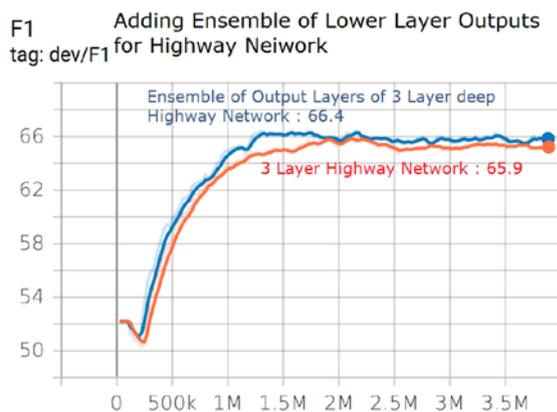

The ensemble of lower layers as output boosted the F1 score from 65.9 to 66.4 for 3 Layer Highway network.

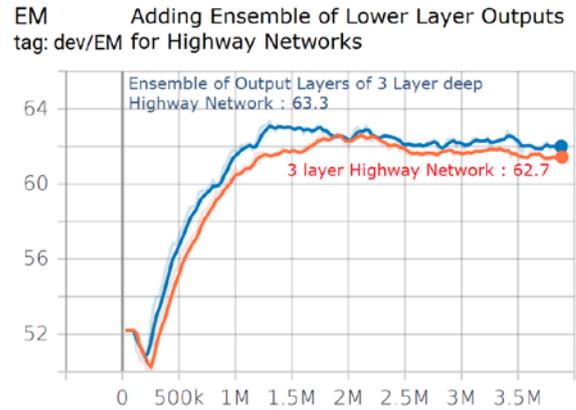

The ensemple of lower layers as output boosted the EM score from 62.7 to 63.3 for 3 Layer Highway network.

### 5.5   BiDAF with addition of character CNN to word level Glove Embedding.

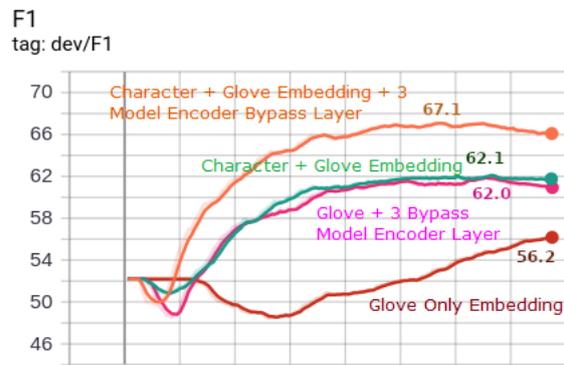

Adding CNN enhanced Character embedding boosted the F1 score from the Glove only word embedding baseline, from 56.2 to 62.1.

Also note that adding the 3 layer deep bypass network of model emcoder boosts the F1 score for Glove only baseline by 6 from 56.2 to 62. In the same way, adding the 3 layer deep bypass network of model emcoder boosts the F1 score for Character + Glove baseline by 5 from 62.1 to 67.1. This proves that the performance boosts in the lower embedding layer is somewhat additive to the performance boosts of the deep networks for model encoders in the higher layer of BiDAF. This is very significant for model optimization as performance boosts becomes additive.

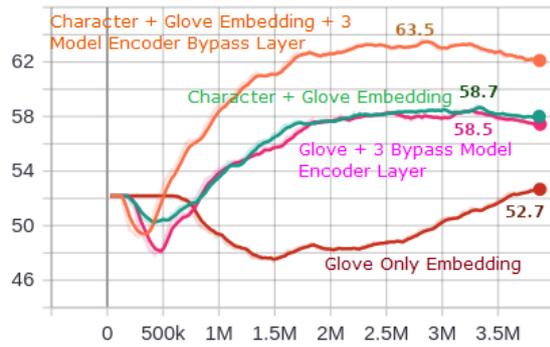

We observe similar performance gains of adding character level embedding to Glove for EM scores. The simple baseline EM score got boosted by 6, from 52.7 to 58.7, and then further about 5 to 63.5 for 3 layer deep bypass network of model encoder.

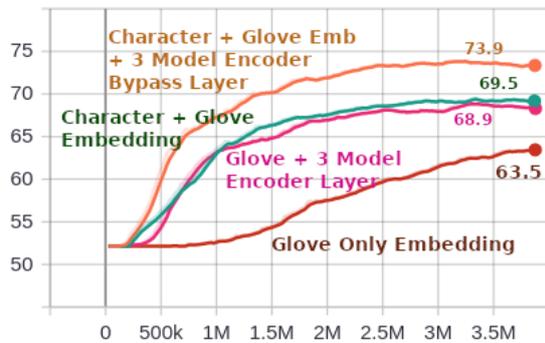

We observe similar performance boosts of character embedding for Answer / No Answer.

### 5.6 Effectiveness of different refinements to BiDAF

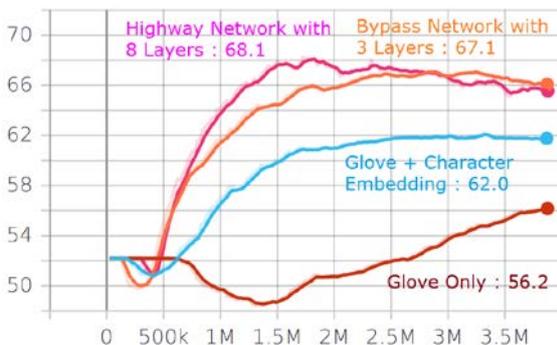

Highway Networks overall performed the best with a F1 score of 68.1 with 8 layers. Bypass network did well, but as the performance dipped after 3 layers, the maximum F1 score it can achieve is 67.1. CNN based Character embedding gave a boost between 5 and 6 of the F1 score for all deep network enhancements to the model encoder layer above it.

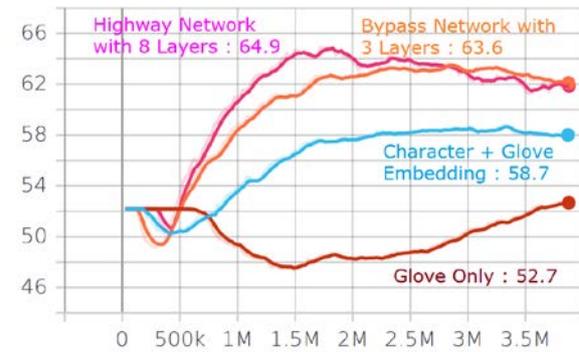

We observe the same trend for EM scores, with highway network of 8 layers topping with 64.9, followed by bypass network with 3 layers at 63.6.

### 6 Analysis

We see significant gains in extension of the architecture of BiDAF with deep highway networks and bypass network for the model encoder layer and in using ensembles of lower layers as output. I did limited experiments of ensembles with multiple layers, and more need to be explored. We also see that adding the CNN character embedding to Glove boosted the performance.

But most significantly, the results show that these refinements at the different layers are additive. As we started with a baseline of word embedding of Glove and then added CNN based character embeddings we see significant gains which are carried forward to the addition of deep networks.

### 7 Conclusion

We tested the hypothesis of whether the deep residual highway networks and its different derivatives are effective in the NLP model, with the model encoder layer of BiDAF as an example, which has both RNN encoder and bi-directional attention flow layers. We observed that both the simple bypass network and the highway network performed very well. With the Highway Network the gains went up till a depth of eight layers, before it went down. This is very significant, as it shows the promise for other NLP model architectures.

The recent trend in NLP has been to go deeper with BERT. Soon the urge to go deeper may be even more insatiable for even higher performance, as applications much more complex than SQUAD question answering have already been deployed. Techniques from the image classification domain like the deep residual networks are bound to come for the deeper NLP networks to make them easily trainable. In this context, this paper's observation of the effectiveness of deep networks is very significant.

Also what is significant is the additive nature of the gains in the embedding layer to the deep networks in the model encoder layer. This further confirms the possibility that as BERT evolves, there is a significant incentive to think beyond a superb language model embedding, and extend the upper layers in handling the complexities of natural language understanding which stretches far beyond the syntactic and immediate context based semantic understanding of language.

## Acknowledgments

I acknowledge the guidance of Christopher Potts and all TAs in the Stanford course.